\pdfoutput=1

\documentclass[11pt]{article}

\usepackage[preprint]{acl}

\usepackage{times}
\usepackage{latexsym}
\usepackage{booktabs}       
\usepackage{multirow}
\usepackage[T1]{fontenc}

\usepackage[utf8]{inputenc}

\usepackage{microtype}

\usepackage{inconsolata}

\usepackage{graphicx}

\title{Large Language Models are Contrastive Reasoners}

\author{Liang Yao \\
  Sun Yat-sen University \\
  Shenzhen, China \\
  \texttt{yaoliang3@mail.sysu.edu.cn} \\
  }

\begin{document}
\maketitle
\begin{abstract}
Prompting methods play a crucial role in enhancing the capabilities of pre-trained large language models (LLMs). We explore how contrastive prompting (CP) significantly improves the ability of large language models to perform complex reasoning. We demonstrate that LLMs are decent contrastive reasoners by simply adding “Let's give a correct and a wrong answer.” before LLMs provide answers. Experiments on various large language models show that zero-shot contrastive prompting improves the performance of standard zero-shot prompting on a range of arithmetic, commonsense, and symbolic reasoning tasks without any hand-crafted few-shot examples, such as increasing the accuracy on GSM8K from $35.9\%$ to $88.8\%$ and AQUA-RAT from $41.3\%$ to $62.2\%$ with the state-of-the-art GPT-4 model. Our method not only surpasses zero-shot CoT and few-shot CoT in most arithmetic and commonsense reasoning tasks but also can seamlessly integrate with existing prompting methods, resulting in improved or comparable results when compared to state-of-the-art methods. Our code is available at the following GitHub repository:~\url{https://github.com/yao8839836/cp}.
\end{abstract}
\section{Introduction}

Recent studies~\cite{zhao2023survey,brown2020language,openai2023gpt4} have shown that large language models (LLMs) exhibit impressive performance across a wide range of tasks. In particular, the chain-of-thought (CoT) prompting technique has demonstrated the capability of LLMs to handle complex tasks, including math problem solving, by guiding them to generate intermediate reasoning steps~\cite{wei2022chain,kojima2022large,zhang2023automatic}. These studies spotlight the significance of developing efficient techniques to direct LLMs in their reasoning processes~\cite{liu2023pre,amatriain2024prompt,chia2023contrastive,yasunaga2023large}.

Nevertheless, the current chain-of-thought (CoT) paradigm encounters two main challenges: offering \textit{correct} guidance or examples of reasoning and reducing the reliance on manual labeling. In particular, Zero-shot-CoT~\cite{kojima2022large} provides general reasoning guidance by providing instructions like “Think step by step.”, but the generated reasoning steps may not be correct and adequate for tasks such as commonsense question-answering (Table~\ref{table:five_runs} and~\ref{tab:answer_extract_csqa_208st}). On the other hand, Few-shot-CoT~\cite{wei2022chain} offers more detailed guidance but necessitates labeled examples of the reasoning process, which can be expensive to obtain for each task. This raises an important research question: Is it possible to generate a more accurate reasoning process without relying on human labeling?

\begin{figure*}[t]
	
	\begin{minipage}{0.5\linewidth}
		\vspace{3pt}

		\centerline{\includegraphics[width=\textwidth]{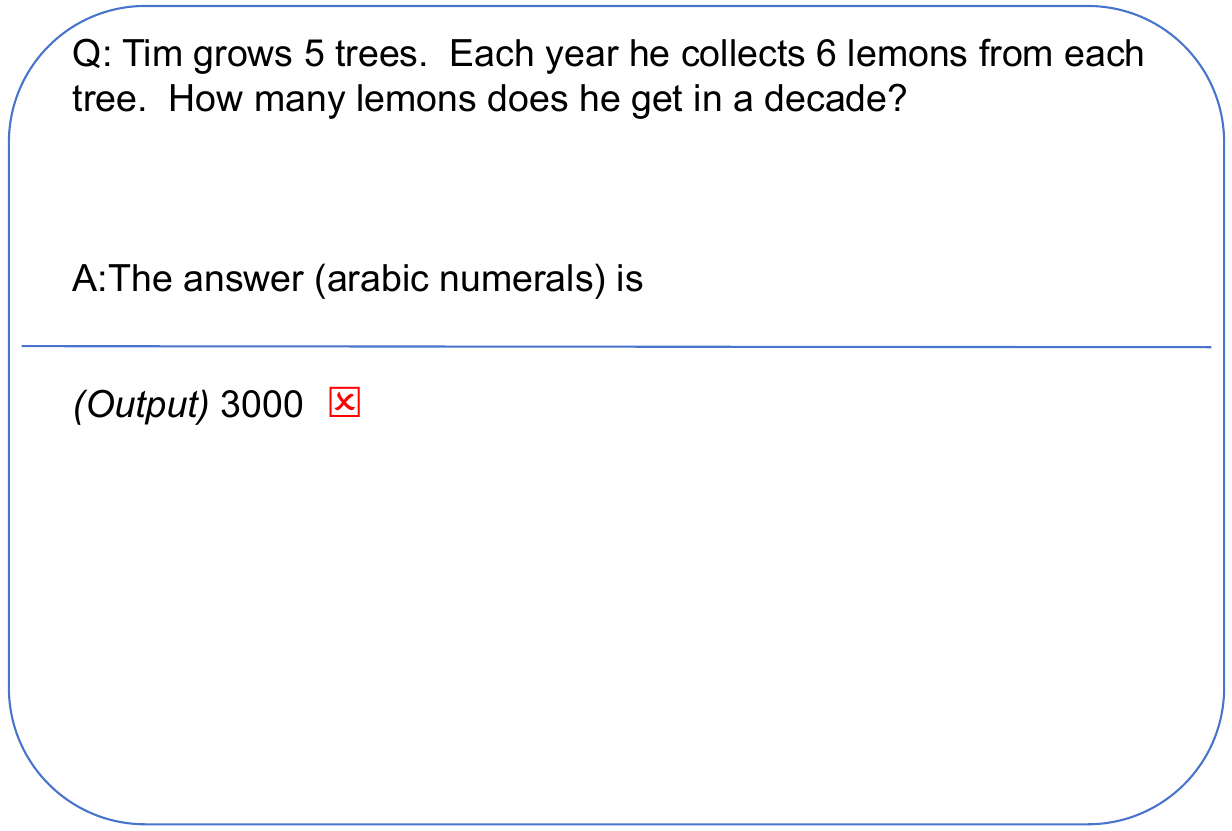}}

		\centerline{(a) Zero-shot}
	\end{minipage}
	\begin{minipage}{0.5\linewidth}
		\vspace{3pt}
		\centerline{\includegraphics[width=\textwidth]{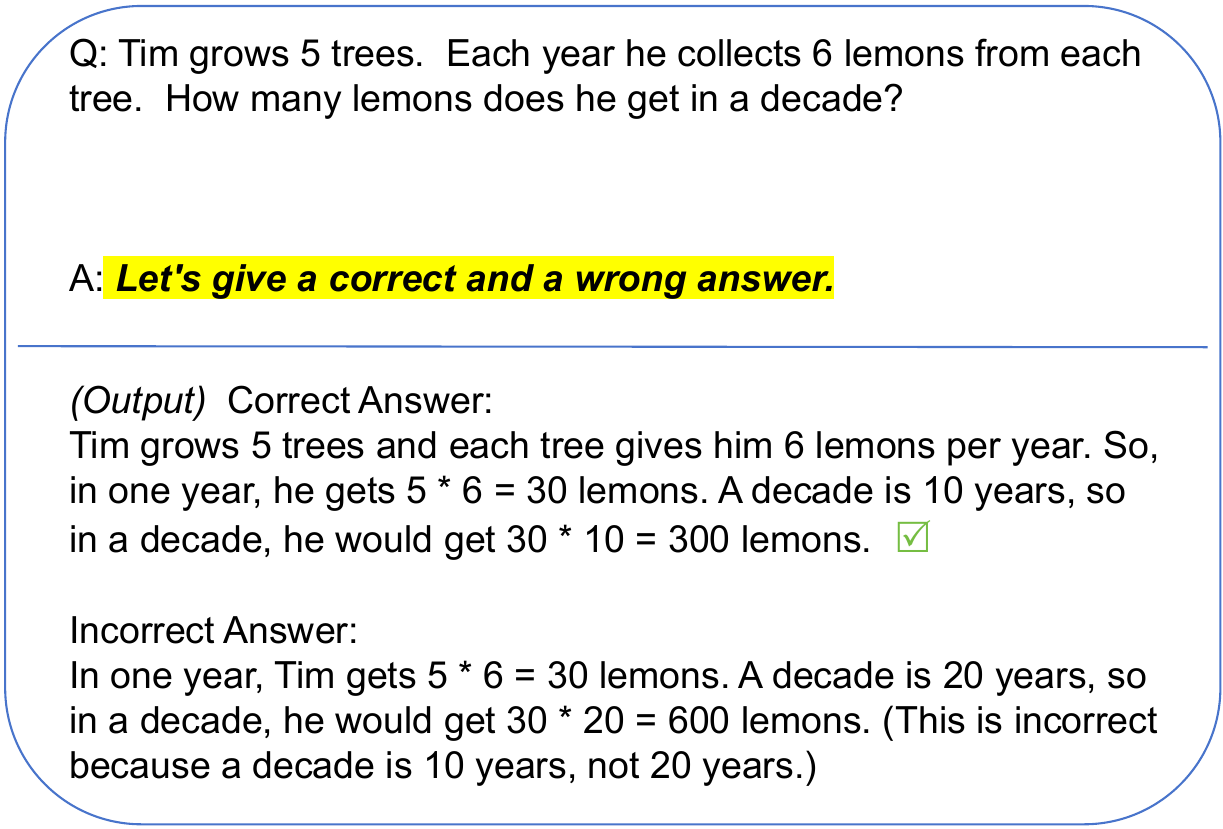}}
	 
		\centerline{(b) Zero-shot-CP (Ours)}
	\end{minipage}
 
	\caption{ Example inputs and outputs of GPT-4 with (a) standard Zero-shot, and (b) ours (Zero-shot-CP). In contrast to Few-shot-CoT, which requires step-by-step reasoning examples for each task, our approach does not rely on any examples. Instead, we use the same prompt "Let's give a correct and a wrong answer" for all tasks, including arithmetic, symbolic, commonsense, and other logical reasoning tasks.}
	\label{fig:zero_shot_cp}
\end{figure*}

In this work, we introduce \textbf{contrastive prompting}, a novel prompting approach that automatically directs
the reasoning process of large language models. Our inspiration stems from how humans can learn from both their correct and incorrect actions~\cite{roediger2009getting}. For instance, when confronted with a math problem (as in Figure~\ref{fig:zero_shot_cp}), people may ask "How can we prevent mistakes in each step?" By identifying the steps that are prone to mistakes on their own, they can enhance their ability to avoid mistakes and provide accurate solutions. Our idea is to prompt LLMs to emulate this reasoning process, enabling them to effectively solve new problems.

Specifically, when presented with a problem to solve, we instruct LLMs to generate both correct and incorrect answers within the given context. To achieve this, we provide prompts such as "Let's give a correct and a wrong answer." Following this, we verify and confirm the correct answer. Our proposed approach offers multiple advantages. It not only generates incorrect answers autonomously but also places a greater emphasis on ensuring the accuracy of the answers. This eliminates the need for manually labeling reasoning examples for each task and problem, effectively addressing the challenges faced by CoT.

We evaluate the proposed approach across various reasoning-intensive tasks, including arithmetic reasoning, commonsense reasoning, symbolic reasoning, and other logical reasoning tasks. We employ two state-of-the-art base LLMs GPT-3.5 and GPT-4~\cite{openai2023gpt4} and four popular open source LLMs. The experimental findings demonstrate significant improvements in scores compared to the zero-shot baseline across all datasets. Moreover, our method not only surpasses Zero-shot-CoT and Few-shot-CoT in most arithmetic and commonsense reasoning tasks but also achieves better results when combined with zero-shot or few-shot CoT, approaching or even surpassing the performance of existing state-of-the-art methods. These results indicate the effectiveness of generating incorrect answers for individual problems to guide the reasoning process of LLMs.


\begin{figure*}[t]
  \centering
  \includegraphics[width = 0.78 \textwidth]{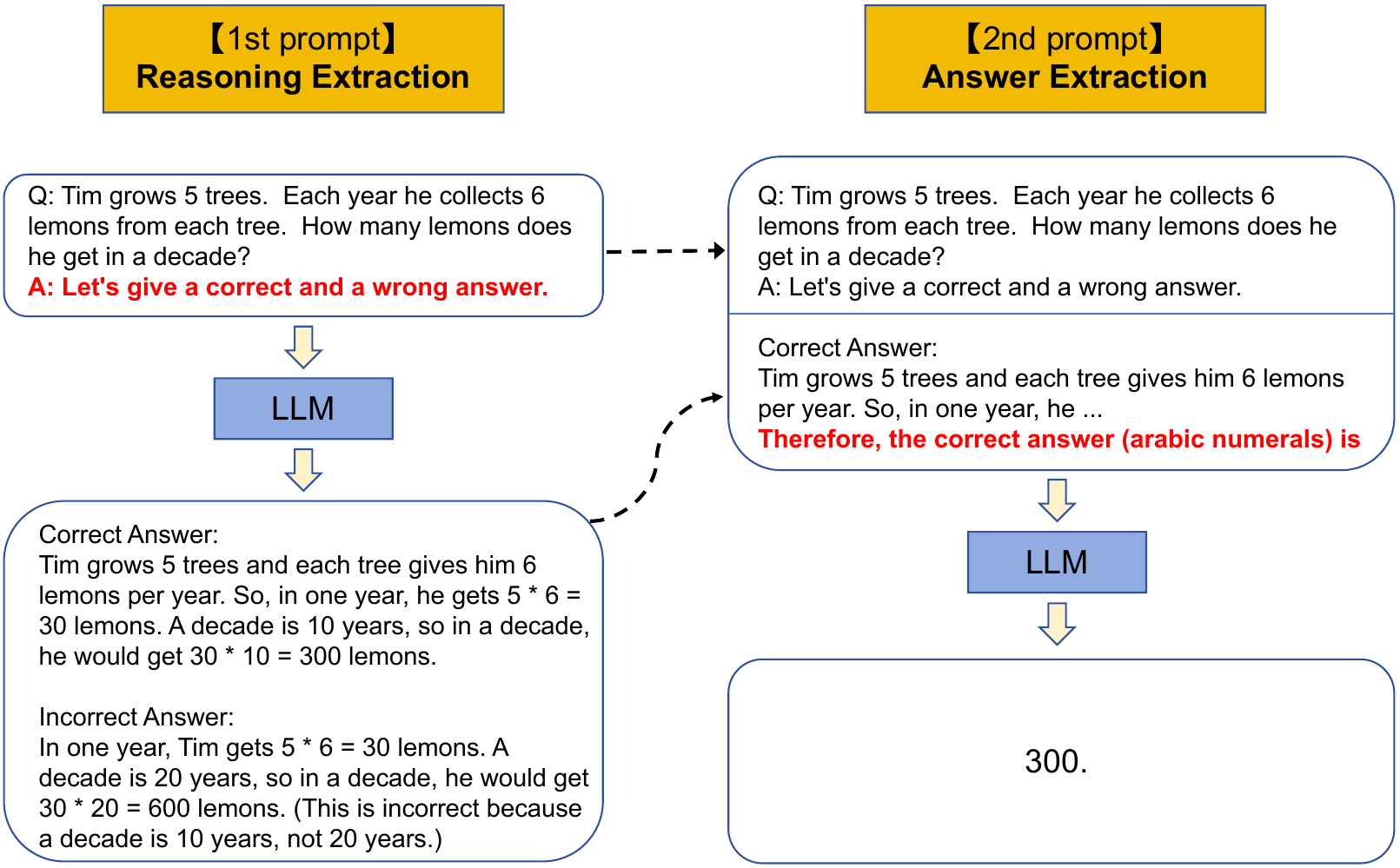}
  \caption{The complete process of Zero-shot-CP involves two steps: Firstly, we utilize the initial "reasoning" prompt to extract a comprehensive reasoning process from a LLM. Secondly, we employ the subsequent "answer" prompt to extract the correct answer from the reasoning text.}
  \label{fig:2steps}
\end{figure*}

\section{Related Works}
\label{related_work}


\paragraph{Large language models and prompting} Recently, LLMs~\cite{zhao2023survey} like ChatGPT and GPT-4~\cite{openai2023gpt4} have gained significant attention. Researchers find that scaling pre-trained language models often leads to an improved model capacity on downstream tasks. These large-sized models show different behaviors from smaller models and display surprising abilities in solving a series of complex tasks.

Prompt engineering is an emerging field dedicated to the development and optimization of prompts, enabling efficient utilization of LLMs across diverse applications and research domains~\cite{amatriain2024prompt,sahoo2024systematic}. Zero-shot prompting involves querying the LLM without any examples while few-shot prompting provides models with a few input-output examples~\cite{brown2020language}. Chain-of-thought (CoT)~\cite{wei2022chain,kojima2022large} prompting enables complex reasoning capabilities through intermediate reasoning steps. Despite its success, Few-shot-CoT~\cite{wei2022chain} needs human-labeled reasoning steps for each example, while Zero-shot-CoT~\cite{kojima2022large} may generate incorrect reasoning steps (especially for commonsense and arithmetic reasoning). Several X-of-thought approaches~\cite{yao2024tree,yao2023beyond,gao2023pal,chen2022program} extend CoT on reasoning tasks, where X can be a tree, a graph, or a program. Auto-CoT~\cite{zhang2023automatic} improves Zero-shot-CoT by providing similar questions as few-shot examples for the target question. Self-consistency~\cite{wang2022self} sample multiple, diverse reasoning paths through Few-shot-CoT, and use the generations to select the most consistent answer. Analogical prompting~\cite{yasunaga2023large} leverages LLMs to automatically generate relevant few-shot examples for each question. In contrast to these works, our method emphasizes eliciting self-awareness in LLMs regarding potential errors and actively avoiding them.

\paragraph{Learning from Negative Examples} Contrastive learning, a widely adopted technique in deep learning, aims to enhance the quality of learned representations by training models to differentiate between "positive" and "negative" samples~\cite{jaiswal2020survey}. In the LLMs area, reinforcement learning from human feedback (RLHF)~\cite{ouyang2022training} and direct preference optimization (DPO)~\cite{rafailov2024direct} fine-tune LLMs with relative human judgments of response quality. Self-reflection~\cite{shinn2024reflexion,kim2024language,madaan2024self,zhang2024context} incorporates "critic" or review steps to identify errors made by the LLM itself and improve upon them. However, it is important to note that the initial output of the LLM may not contain any errors, and there is a potential risk of the model reinforcing its own errors if it inaccurately evaluates the quality of its responses or generates invalid principles. The closest work to ours is the Contrastive CoT~\cite{chia2023contrastive} that extends Few-shot-CoT by creating wrong reasoning processes from annotated correct reasoning steps. The main distinction is that the erroneous answers generated by Contrastive CoT still require human-annotated examples, and the random reordering of entities during the reasoning process may not align with the patterns of errors made by LLMs themselves. On the contrary, our approach enables LLMs to generate erroneous answers on their own, which aligns better with their intrinsic knowledge. It does not require human annotation.

\section{Contrastive Prompting}
\label{method}
We propose Contrastive Prompting (CP), a template-based prompting approach for contrastive reasoning. Our method can seamlessly integrate with any prompting technique by incorporating a trigger sentence before the LLM provides answers. In the following, we first illustrate our method using Zero-shot-CP as an example, which only uses the original question without supporting examples. Next, we will discuss how to combine our method with other prompting techniques.


\subsection{Two-stage prompting}
Although Zero-shot-CP is straightforward in concept, it utilizes prompting twice to extract both reasoning and answer, as illustrated in Figure~\ref{fig:2steps}.

\paragraph{1st prompt: reasoning extraction} In this step we begin by transforming the input question $\mathbf{x}$ into a prompt $\mathbf{x'}$ using a simple template “Q: [X]. A: [T]”. Here [X] represents the input slot for $\mathbf{x}$ and [T] represents a slot for a manually crafted trigger sentence $\mathbf{t}$ that would extract the reasoning process to answer the question $\mathbf{x}$. For instance, if we use “Let's give a correct and a wrong answer.” as a trigger sentence, the prompt $\mathbf{x'}$ would be “Q: [X]. A: Let's give a correct and a wrong answer.”. Additional trigger examples can be found in Table~\ref{table:templates}. Prompted text $\mathbf{x'}$ is then inputted into a LLM, which generates the subsequent sentence $\mathbf{z}$. 


\paragraph{2nd prompt: answer extraction} In the second step, we utilize the generated sentence $\mathbf{z}$ in conjunction with the prompted sentence $\mathbf{x'}$ to extract the ultimate answer from the LLM. To provide a more specific explanation, we combine three elements by concatenating them as "[X'] [Z] [A]". Here, [X'] represents the 1st prompt $\mathbf{x'}$, [Z] represents the sentence $\mathbf{z}$ generated in the first step, and [A] represents a trigger sentence used to extract the answer. The prompt for this step is self-augmented, meaning that it includes the sentence $\mathbf{z}$ generated by the same LLM. During the experiment, we employed slightly different answer triggers based on the format of the answer. Please refer to Appendix~\ref{appendix:answer_extract} for the answer trigger sentences we used in each task. Subsequently, the prompted text is inputted into the LLM to generate sentences $\mathbf{y}$ and extract the final answer. 


\subsection{Integrating with other prompting methods}
We can easily integrate our CP with any advanced prompting methods. We name the combined method X-CP, where X can be Zero-shot-CoT, Few-shot-CoT, or any other method. X-CP also has two steps: reasoning extraction and answer extraction. For Zero-shot-CoT-CP, the only distinction is we replace the trigger sentence “Let's give a correct and a wrong answer.” with “Let's think step by step and give both a correct answer and a wrong answer.”. For Few-shot-CoT-CP, the distinction is that $k$ few-shot examples with reasoning steps are added before “Q: [X]. A: Let's give a correct and a wrong answer.”, the resulting prompt $\mathbf{x'}$ will be “Q: [$X_1$] A: [$Z_1$]. The answer is [$Y_1$]. Q: [$X_2$] A: [$Z_2$]. The answer is [$Y_2$]. ... Q: [$X_k$] A: [$Z_k$]. The answer is [$Y_k$]. Q: [X]. A: Let's give a correct and a wrong answer.”, where $X_i$, $Z_i$ and $Y_i$ are the question, reasoning steps and the final answer for each example $i$.

\section{Experiment}
\label{exp}

\subsection{Settings}
\paragraph{Datasets} We evaluate the effectiveness of our proposal on 12 datasets\footnote{The datasets are available at~\url{https://github.com/kojima-takeshi188/zero_shot_cot/tree/main/dataset}.} encompassing four categories of reasoning tasks: arithmetic (SingleEq, AddSub, MultiArith, AQUA-RAT, GSM8K, SVAMP), commonsense (CommonsenseQA, StrategyQA), symbolic (Last Letter Concatenation, Coin Flip), and other logical reasoning tasks (Date Understanding, Tracking Shuffled Objects). The detailed description of each dataset can be found in~\cite{kojima2022large}. We use the few-shot examples with reasoning steps provided by~\cite{wei2022chain}.


\paragraph{Baselines} We conducted a comprehensive comparison of our CP method with various types of prompting techniques. These include simple zero-shot methods such as Zero-shot and Zero-shot-CoT~\cite{kojima2022large}, Few-shot and Few-shot-CoT~\cite{wei2022chain}, X-of-thought approaches like Tree of Thoughts (ToT)~\cite{yao2024tree}, Graph of Thoughts (GoT)~\cite{yao2023beyond}, Program-aided Language models (PAL)~\cite{gao2023pal}, and Program of thoughts prompting (PoT)~\cite{chen2022program}. Additionally, we compared our method with other prompting techniques such as Analogical prompting (Self-generated Exemplars)~\cite{yasunaga2023large} and Self-consistency (SC)~\cite{wang2022self}. Furthermore, we evaluated the effectiveness of self-reflection methods, including Recursive Criticism and Improvement (RCI)~\cite{kim2024language}, Self-Refine~\cite{madaan2024self} and Learning Principles from Mistakes (LEAP)~\cite{zhang2024context}, as well as the closest related work, Contrastive CoT~\cite{chia2023contrastive}.  We also experimented with running CP using Self-consistency (SC). Specifically, we set the temperature parameter of LLMs to 0.7 and sampled 10 correct and incorrect answers. Then, we selected the answer that appeared most frequently among the 10 correct answers as the final answer.

\paragraph{Models} We use GPT-4 and GPT-3.5-Turbo (0613) as our base models (accessed between Feb 22nd–May 22nd 2024) for main experiments. We also tested our CP on various open LLMs: LLaMA3-8B, LLaMA3-70B~\cite{touvron2023llama}, ChatGLM3-6B~\cite{du2022glm} and Qwen1.5-72B-Chat~\cite{qwen}. All generations (except experiments with Self-consistency) are done by greedy decoding (i.e., sampling with zero temperature) as in the original CoT work~\cite{wei2022chain}. For GPT models, we use Azure OpenAI services. For open LLMs except ChatGLM3-6B, we use LlamaAPI~\footnote{\url{https://docs.llama-api.com/quickstart}}. For ChatGLM3-6B, we downloaded the model and performed the inference on a Linux server with an A100 GPU. 

\paragraph{Answer filtering} We follow Zero-shot-CoT~\cite{kojima2022large} work and use its original implementation to pick up the final answers.

\begin{table*}[t]
  \footnotesize
  \centering
  \begin{tabular}{lcccccc}
    \toprule
        & MultiArith    & GSM8K & StrategyQA & AddSub & SVAMP & CommonsenseQA\\
    \midrule
    Zero-shot &  60.97 & 14.39 & 65.02& 82.78& 69.74 &71.33\\
    Zero-shot-CoT & 94.87 & \textbf{75.56} & 60.74& 86.16& 81.78 &68.96\\
    Zero-shot-CP & \textbf{95.13} & 73.22 & \textbf{67.39}& \textbf{90.46}& \textbf{83.08} & \textbf{73.81}\\
    \bottomrule
  \end{tabular}
    \caption{Accuracy (in percentage) comparison of Zero-shot-CP with Zero-shot and Zero-shot-CoT on five datasets. We run all methods 5 times using GPT-3.5-Turbo and report average scores. Zero-shot-CP significantly outperforms baselines on StrategyQA, AddSub, SVAMP and CommonsenseQA based on student t-test (p < 0.05).}
    \label{table:five_runs}
\end{table*}

\subsection{Results}
\paragraph{Zero-shot Results} 
Table~\ref{table:five_runs} presents the accuracy scores achieved by our Zero-shot-CP, standard zero-shot prompting (Zero-shot) and Zero-shot-CoT across five datasets. We ran all methods five times using GPT-3.5-Turbo and report the average scores. We found that the differences in each run were minimal. Zero-shot-CP consistently outperformed Zero-shot-CoT and Zero-shot across most (4 out of 5) datasets.

Table~\ref{table:main_results_zero} in Appendix~\ref{appendix:extra_results} presents more comprehensive results. Notably, Zero-shot-CP demonstrates significant improvements over Zero-shot on all 12 datasets across various tasks using GPT-3.5-Turbo. For instance, Zero-shot-CP achieves score gains ranging from $14.3\%$ to $73.2\%$ on GSM8K, from $61.2\%$ to $95.2\%$ on MultiArith and from $4.2\%$ to $41.8\%$ on Last Letter Concatenation. Moreover, Zero-shot-CP outperforms Zero-shot on the majority (9 out of 12) of datasets when using GPT-4, with improvements ranging from $35.9\%$ to $88.8\%$ on GSM8K and from $41.3\%$ to $62.2\%$ on AQUA-RAT. These results indicate that eliciting self-awareness in LLMs to compare incorrect and correct answers can help prevent incorrect responses.

Zero-shot-CP outperforms Zero-shot-CoT in the majority (4 out of 6) of arithmetic reasoning tasks, suggesting that the self-awareness of LLMs regarding incorrect answers may be more crucial than their self-awareness regarding steps in mathematical reasoning. Furthermore, in commonsense reasoning tasks, Zero-shot-CP consistently outperforms Zero-shot (2 out of 2), while Zero-shot-CoT exhibits inferior results. This is likely because commonsense reasoning tasks require fewer steps, making awareness of individual pieces of commonsense knowledge more crucial. However, Zero-shot-CP performs worse than Zero-shot-CoT in symbolic reasoning and other reasoning tasks, indicating that for tasks like Coin Flip and Shuffled Objects, the steps involved are more crucial. Moreover, breaking down the reasoning process into individual steps makes it less likely for LLMs to make mistakes in these tasks, as the action space is very limited. 
Nevertheless, by combining Zero-shot-CP and Zero-shot-CoT, Zero-shot-CoT-CP achieves better results in most tasks (see Table ~\ref{table:main_results_baselines}). 

Table~\ref{table:results_openllm} in Appendix~\ref{appendix:extra_results} presents the results of using various open-source LLMs: LLaMA3-8B, LLaMA3-70B, ChatGLM3-6B, and Qwen1.5-72B-Chat as base models. The results demonstrate that Zero-shot-CP not only performs well with state-of-the-art GPT models, but also exhibits significant improvements across multiple sizes of open-source models.

\begin{table*}[t]
\scriptsize

  \centering
  \begin{tabular}{l|l}
    \toprule
    \multirow{3}{*}{Question}   & Q: Danny collects bottle caps and wrappers. He found 46 wrappers and 50 bottle caps at the park.\\
                  &~~~~~Now he has 21 bottle caps and 52 wrappers in his collection.\\ 
                  &~~~~~How many more bottle caps than wrappers did danny find at the park?\\
    \midrule
    \multirow{2}{*}{Zero-shot} & A: The answer (arabic numerals) is {\color{red}-4}. Danny found 50 bottle caps and 46 wrappers at the park,\\ 
    &~~~~~so he found 4 more bottle caps than wrappers.   \\
    \midrule
    \multirow{8}{*}{Zero-shot-CoT} & A: Let's think step by step. First, we need to find out how many bottle caps and wrappers\\ 
    &~~~~~Danny had before he found the new ones at the park.   \\
    &~~~~~We know that after finding the new ones, he has 50 bottle caps and 52 wrappers.\\ 
    &~~~~~Before finding the new ones, he had 21 bottle caps and 46 wrappers.\\ 
    &~~~~~So, the number of bottle caps he found at the park is 50 (total now) - 21 (total before) = 29 bottle caps.\\ 
    &~~~~~The number of wrappers he found at the park is 52 (total now) - 46 (total before) = 6 wrappers.\\
    &~~~~~Therefore, Danny found 29 - 6 = 23 more bottle caps than wrappers at the park. \\
    &~~~~~Therefore, the answer (arabic numerals) is {\color{red}23} \\ 
    \midrule
    \multirow{6}{*}{Zero-shot-CP} & A: Let's give a correct and a wrong answer. \textbf{Correct Answer}: Danny found 50 bottle caps and 46 wrappers\\ 
    &~~~~~at the park. So, he found 50 - 46 = 4 more bottle caps than wrappers at the park. \\
    & \\
    &~~~~~\textbf{Incorrect Answer}: Danny found 50 bottle caps and 46 wrappers at the park.\\
    &~~~~~So, he found 46 - 50 = -4 more bottle caps than wrappers at the park. \\
    &~~~~~Therefore, the \textbf{correct} answer (arabic numerals) is {\color{green}4} \\    
    \midrule
  Ground Truth  &~~~~~4  \\  
    \bottomrule
  \end{tabular}
  \caption{An example question from SVAMP and responses from Zero-shot, Zero-shot-CoT, Zero-shot-CP and Zero-shot-CoT-CP. We use GPT-4 as the base model.}
    \label{tab:answer_extract_svamp_42st}
\end{table*}

\paragraph{Qualitative Analysis} Table~\ref{tab:answer_extract_svamp_42st} and Table~\ref{tab:answer_extract_csqa_208st} in Appendix ~\ref{appendix:extra_results} illustrates examples from SVAMP and CommonsenseQA. For the example from the arithmetic reasoning task SVAMP, we found that the reasoning process of zero-shot is correct, but it produces an incorrect answer "-4". Zero-shot-CoT is disrupted by irrelevant information, resulting in incorrect reasoning processes and answers being generated. Zero-shot-CP, on the other hand, is not disrupted and provides both the correct answer and explanation. We can see that the "wrong answer" "-4" from Zero-shot-CP is a real mistake made by Zero-shot. For the example from the common sense reasoning task CommonsenseQA, contrastive prompting is able to recognize the word "work" in the question and provide the correct answer, while Zero-shot and Zero-shot-CoT cannot. 

In Appendix~\ref{appendix:extra_results}, we present responses generated by Zero-shot-CP for each dataset. Figure~\ref{fig:example_addsub}--\ref{fig:example_svamp} gives both a positive example and a negative example of Zero-shot-CP on each dataset. From positive examples, we found that Zero-shot-CP can generate "wrong" answers that are indeed incorrect in most cases (11/12), except for Tracking Shuffled Object (Figure~\ref{fig:example_object}). Incorrect answers are generated by intentionally calculating inaccurately (Figure~\ref{fig:example_multiarith}), disregarding important details (Figure~\ref{fig:example_gsm8k}), searching for descriptions that are not present in the question (Figure~\ref{fig:example_csqa}), or deliberately providing explanations that contradict common sense (Figure~\ref{fig:example_strategyqa}). From negative examples, We found that the "wrong answers" provided by Zero-shot-CP can actually be valid answers (Figure~\ref{fig:example_aqua}, \ref{fig:example_date}, \ref{fig:example_coinflip}, \ref{fig:example_multiarith}, \ref{fig:example_singleeq} and \ref{fig:example_strategyqa}). In some other negative examples, both the "correct answers" and "incorrect answers" provided by Zero-shot-CP are inconsistent with the ground truth (Figure~\ref{fig:example_addsub}, \ref{fig:example_csqa}, \ref{fig:example_gsm8k}, \ref{fig:example_lastletter} and \ref{fig:example_svamp}). From the figures, we found that Zero-shot-CP also outputs reasoning steps in the process of generating correct and incorrect answers, especially for arithmetic reasoning tasks. Furthermore, we manually annotated 10 solved problems and 10 unsolved problems of Zero-shot-CP for each of the 12 datasets. Table~\ref{table:240_examples} provides the categorization and counts of these 120 solved problems and 120 unsolved problems. We found that for the solved problems, the majority (112/120) of the given "wrong" answers were indeed incorrect. For the unsolved problems, the majority (91/120) of both the "correct" and "wrong" answers were incorrect, with a portion (23/120) of the "wrong" answers actually being the ground truth. This situation typically occurs in yes or no questions.

\begin{table*}[t]

  \scriptsize
  \centering
  \begin{tabular}{lcccc}
    \toprule
     GPT-4   & AQUA    & GSM8K & AddSub & MultiArith  \\
    \midrule
    Let's give a correct and a wrong answer. &  62.2 & 88.8 & \underline{\textbf{91.6}} & \underline{\textbf{97.8}}\\
    Let's first give a wrong answer, then give the correct answer. & 69.3 & 86.1 & 90.9& 95.0\\
    Let's first give the correct answer, then give a wrong answer. & 58.7 & \textbf{89.7} & 91.6 & 95.0 \\
    Let's give a correct and an incorrect answer.  &  66.5 & 88.7 &  91.6& 97.7\\
    Please give a correct and a wrong answer. & 57.5 & 82.0 & 88.9 & 94.0\\
    Let's give a correct answer. & \underline{\textbf{71.7}} & 75.9 & 89.4 & 97.0\\
    \midrule
    Let's think step by step and give both a correct answer and a wrong answer. & \textbf{71.3} & 89.5 & \textbf{91.4} & 97.2\\
    Let's give a correct and a wrong answer. Let's also think & \multirow{2}{*}{52.8} & \multirow{2}{*}{88.9} & \multirow{2}{*}{89.4} & \multirow{2}{*}{96.7} \\
    step by step for the correct and the wrong answer. &  &  &  & \\
    Let's think step by step. (Zero-shot-CoT) & 70.1 & \underline{\textbf{90.9}} & 89.6& \textbf{97.7}\\
    \bottomrule
  \end{tabular}
  \caption{Comparison prompting templates using accuracies (in percentage) on AQUA-RAT, GSM8K, AddSub and MultiArith in zero-shot setting. GPT-4 is used as the model. Bolded numbers indicate the best results within each block's column, while underlined numbers indicate the best results across the entire column.}
  \label{table:templates}
\end{table*}

\begin{table*}[t]
  
  \scriptsize
  \centering
  \begin{tabular}{lccccc}
    \toprule
        & MultiArith    & GSM8K & StrategyQA & AQUA & SVAMP \\
    \midrule
    Zero-shot &  61.2 & 14.3 & 65.0& 29.9& 69.7\\
    Zero-shot-CoT & 94.8 & 75.1 & 60.9& 55.9& 81.9\\
    Zero-shot-CoT + SC & 96.8 & \textbf{80.7}& 61.6& \textbf{66.1}& 85.6\\
    Zero-shot-CP & 95.2 & 73.2 & 67.3& 40.2& 83.2\\
    Zero-shot-CP + SC & \textbf{98.3} & 80.3 & \textbf{67.9}& 48.4& \textbf{87.6}\\
    Zero-shot-CoT-CP & 96.2 & 73.5 & 66.7 & 60.6& 85.9\\
    \midrule
    Few-shot &  87.3& 58.2 & 56.7& 37.4& 78.2\\
    Few-shot-CoT &  98.0& 71.1 & 62.2& 55.5& 81.0\\
    Few-shot-CoT + SC & 98.7 & 76.0 & 63.5& 59.4& 83.5\\
    Few-shot-CoT-CP & 97.5 & 72.7 &68.7 &52.0& 82.2\\    
    Few-shot-CoT (GPT-4) &  98.3& 89.5 & \textbf{79.1}& 58.7& 83.3\\    
    Few-shot-CoT-CP (GPT-4) & \textbf{98.7} & 90.3 &78.2 &66.9& 91.8\\
    Few-shot-CoT-CP (GPT-4) + SC & 97.5 & \underline{\textbf{91.9}} &78.8 & \underline{\textbf{70.9}}& \underline{\textbf{93.1}}\\
    Contrastive CoT~\cite{chia2023contrastive} & -- &  79.0& 66.2 &57.5& 81.6\\
    \midrule
    Self-consistency (Code-davinci-002)~\cite{wang2022self}& \underline{\textbf{100.0}} & 78.0 & 79.8 & 52.0& 86.8\\
    PAL (Codex)~\cite{gao2023pal}& 99.2 & 80.4 & -- & --& 79.4\\
    Zero-shot-PoT (Codex)~\cite{chen2022program}& 92.2 & 57.0 & -- & 43.9 & 70.8\\
    Few-shot-PoT (Codex)~\cite{chen2022program}& -- & 71.6 & -- & 54.1 & 85.2\\
    Few-shot-PoT-SC (Codex)~\cite{chen2022program}& -- & 80.0 & -- & \textbf{58.6} & \textbf{89.1}\\
    ToT (GPT-4)~\cite{yao2024tree}& -- & \textbf{90.0} & \underline{\textbf{83.0}}& --&  --\\
    GoT (T5-large)~\cite{yao2023beyond}& -- & 82.2 & --& --&  --\\
    Self-generated Exemplars~\cite{yasunaga2023large}& -- & 77.8 &-- &--&-- \\
    Self-Refine~\cite{madaan2024self}& -- & 75.1 &-- &--&-- \\
    LEAP~\cite{zhang2024context}& -- & 77.4 &-- &--&-- \\
    Zero-Shot-CoT + RCI ~\cite{kim2024language}& 97.2 & 86.2 &-- & --& 85.8 \\
    Few-Shot-CoT + RCI ~\cite{kim2024language}& 99.2 & 84.3 &-- & --& 87.4 \\
    \bottomrule
  \end{tabular}
    \caption{Comparison with baseline methods using accuracies (in percentage) on MultiArith, GSM8K, StrategyQA, AQUA-RAT and SVAMP. GPT-3.5-Turbo is used as the model if not specified. The baseline results with citations are obtained from corresponding papers. Bolded numbers indicate the best results within each block's column, while underlined numbers indicate the best results across the entire column.}
    \label{table:main_results_baselines}
\end{table*}


\paragraph{The impact of prompt selection on Zero-shot-CP} We explore different contrastive prompts and their combination with Zero-shot-CoT. Table~\ref{table:templates} outlines performance using 9 different templates with two classes. The first category is related to correct and wrong answers. We found "Let’s give a correct and a wrong answer." achieves the best results in general. "Let's first give a wrong answer, then give the correct answer." performs well on AQUA-RAT but it performs worse on other datasets. "Let's first give the correct answer, then give a wrong answer." generally performs well on the four datasets, meaning that providing the correct answer first and then the incorrect answer generally leads to better results. The trigger word "incorrect" performs similarly to "wrong", and the trigger word "Please" performs much worse than "Let's". This is likely because, in the pre-training and fine-tuning data, there are slightly fewer occurrences of "incorrect" compared to "wrong" in samples related to correct and incorrect answers, and "Please" is rarely present as this type of data is generally not dialogue data. "Let's give a correct answer." performs well on the multiple-choice question dataset AQUA-RAT, but the performances on other three mathematical reasoning tasks are not satisfactory. This indicates that, for multiple-choice questions, only providing a correct answer is equivalent to eliminating several incorrect answers. However, for questions without options, outputting an incorrect answer is helpful. 

Table~\ref{table:results_correct_only} in Appendix~\ref{appendix:extra_results} gives more comparative results between "Let's give a correct and a wrong answer." and "Let's give a correct answer." We find that, except for multiple-choice reasoning tasks, providing a wrong answer is more effective than only giving the correct answer.  We also printed the token output probabilities for different prompts. As shown in Figure~\ref{fig:token_prob} in Appendix~\ref{appendix:extra_results}, we find that adding prompts to generate incorrect answers changes the output probability distribution, Zero-shot-CP makes GPT-4 more confident in the ground truth answer.

The second type of template in Table~\ref{table:templates} connects to Zero-shot-CoT, and we found that starting with the steps performs better than starting with the correct and wrong answers. Overall, it appears that Zero-shot-CoT-CP ("Let's think step by step and give both a correct answer and a wrong answer.") performs the best.

\begin{figure}[t]
	\begin{minipage}{0.24\linewidth}
		\vspace{3pt}
		\centerline{\includegraphics[width=\textwidth]{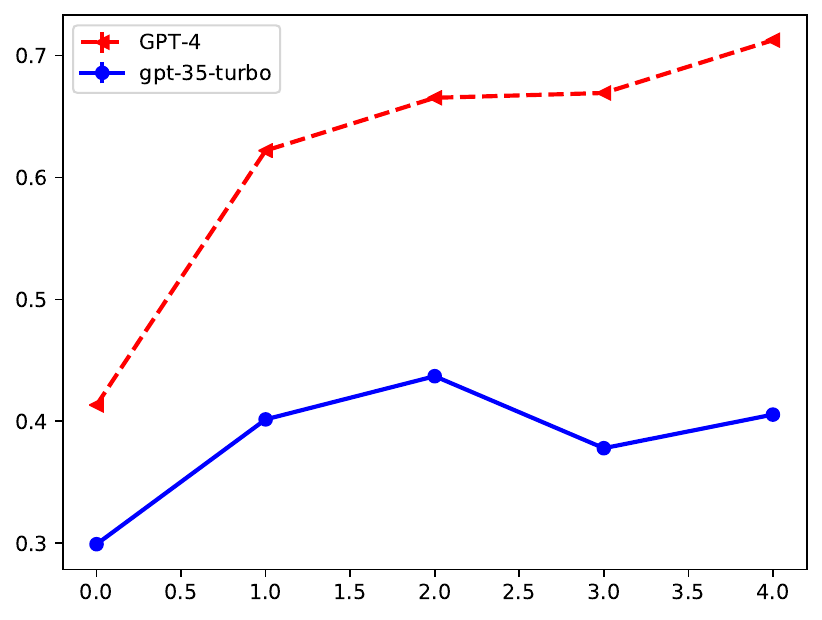}}
		\centerline{AQUA}
	\end{minipage}
	\begin{minipage}{0.24\linewidth}
		\vspace{3pt}
		\centerline{\includegraphics[width=\textwidth]{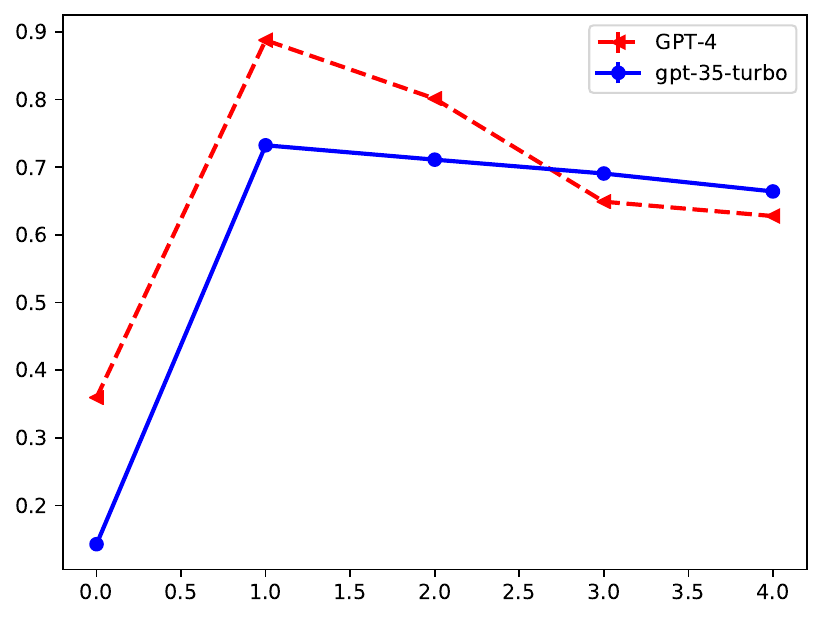}}
		\centerline{GSM8K}
	\end{minipage}
    \begin{minipage}{0.24\linewidth}
		\vspace{3pt}
		\centerline{\includegraphics[width=\textwidth]{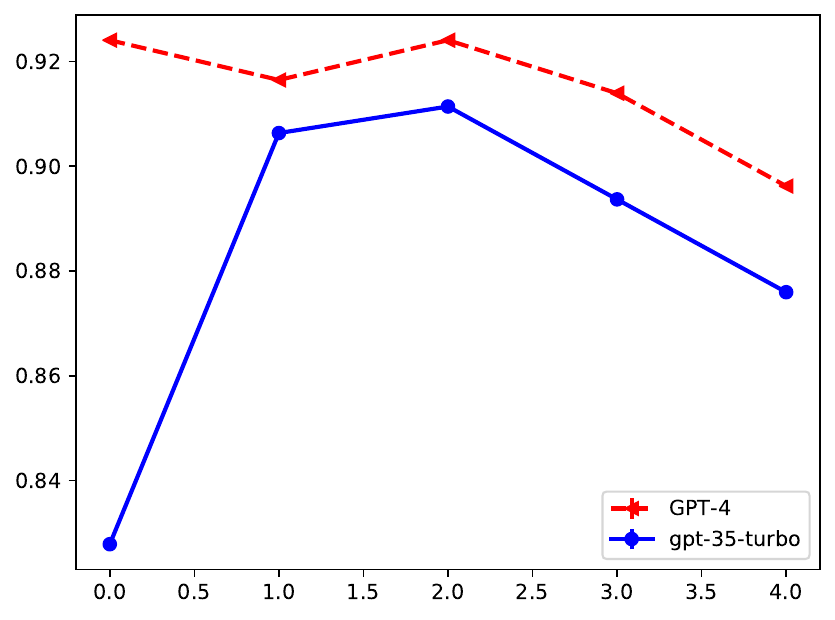}}
		\centerline{AddSub}
	\end{minipage}
    \begin{minipage}{0.24\linewidth}
		\vspace{3pt}
		\centerline{\includegraphics[width=\textwidth]{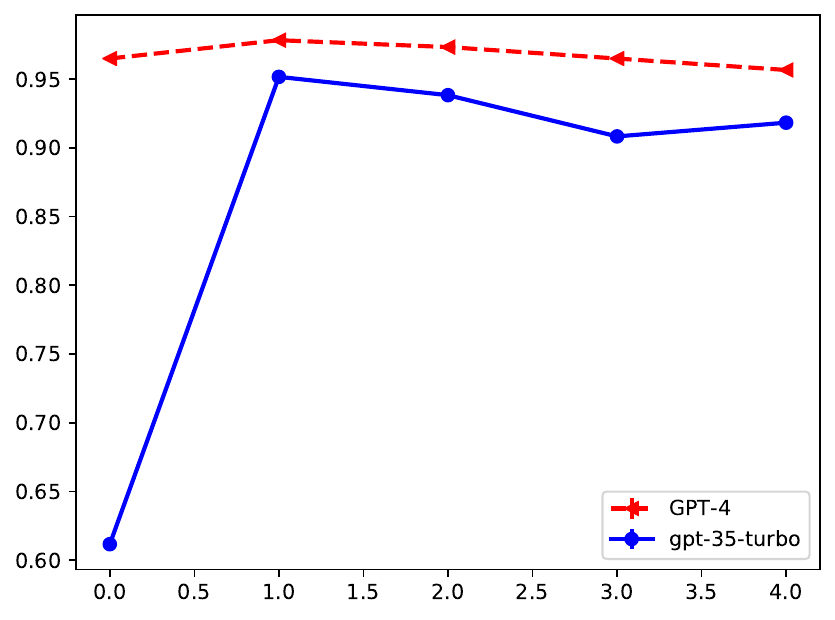}}
		\centerline{MultiArith}
	\end{minipage}
	\caption{ Accuracy scores by varying the number of wrong answers. We test GPT-4 and GPT-3.5-Turbo on (a) AQUA-RAT, (b) GSM8K, (c) AddSub and (d) MultiArith. The range of the number of wrong answers is from 0 (Zero-shot) to 4.}
	\label{fig:num_wrong}
\end{figure}

\paragraph{The impact of number of wrong answers on Zero-shot-CP} We explored the impact of the number of incorrect answers on accuracy. We vary the number of wrong answers from $0$ to $4$, where $0$ means standard zero-shot prompting. For $k = 1,2,3,4$, we use the template "Let's give a correct and $k'$ wrong answer(s).", where $k'$ can be "a", "two", "three" and "four". Figure~\ref{fig:num_wrong} plots the results. We found that providing 1-2 incorrect answers yielded the best results in general. The only exception is on AQUA-RAT, where providing more incorrect answers resulted in higher accuracy. This is because the task involves a multiple-choice question with five options, and excluding more incorrect answers makes the LLMs more certain about the correct answer. For math reasoning tasks with an infinite number of answers, providing just one incorrect answer seems to be sufficient.

\paragraph{Comparison with other baselines} Table~\ref{table:main_results_baselines} compares the performances on four mathematical reasoning datasets (MultiArith, GSM8K, AQUA-RAT and SVAMP) and one common sense reasoning dataset (StrategyQA) across CP and baselines. We find that Zero-shot-CP not only outperforms Few-shot, but also achieves comparable or even superior results to Few-shot-CoT. For instance, on GSM8K, the absolute accuracy has improved by $2.1\%$, and on StrategyQA, the absolute accuracy has improved by $5.1\%$. This suggests that in certain cases, the provided examples and reasoning steps may not be as effective as directly triggering the LLM's self-awareness of errors. By combining CP and Few-shot-CoT, we can achieve even better results. Furthermore, if we utilize the GPT-4 model, we can attain performance that is comparable to or even superior to the current state-of-the-art methods. For example, in AQUA, SVAMP, and GSM8K, we have achieved higher accuracy scores compared to recently published results. When running CP with Self-consistency (SC), the scores can be further improved in both zero-shot and few-shot settings.

For a more in-depth performance analysis, we note that X-of-thought methods can improve the effectiveness of Few-shot-CoT, indicating that trees, graphs, and code indeed provide richer information and greater flexibility compared to simple chains of thought. Among them, the results reported by the ToT work seem to be more prominent. By sampling multiple reasoning paths and selecting the most consistent answer, Self-consistency (SC) demonstrates excellent performance in mathematical and commonsense reasoning tasks. It can also be effectively combined with other methods such as PoT. Self-generated Exemplars also show better performance than CoT, indicating that allowing the LLM to recall relevant questions and answer them before responding to the original question is helpful. The performance of Self-reflection methods, such as Self-Refine and LEAP, is similar to that of Self-generated Exemplars. RCI performs even better, primarily due to its direct combination with the CoT method. Compared to these methods, our approach is simpler and can also yield comparable results. Compared to the most relevant method Contrastive CoT, our Zero-shot-CP performs better on the StrategyQA and SVAMP datasets. Zero-shot-CoT-CP performs better on AQUA-RAT. However, on GSM8K, Contrastive CoT performs better, indicating that generating incorrect answers by swapping the order of entities is useful for this task.


The main reasons why CP works well are fourfold: 1) the pre-training data of LLMs contains many correct and incorrect answers to different types of questions. For instance, many web pages and books in Appendix~\ref{appendix:pretrain_data_examples} provide correct and incorrect answers to math reasoning and common sense reasoning questions. Answers to questions on social media platforms like Reddit, Quora, and Zhihu can be voted on by others through “upvotes” or “downvotes”. Highly upvoted answers are more likely to be correct answers while others may be incorrect. Pre-training LLMs with massive text containing these correct and wrong answers can encode general patterns (token probability) of these answers in LLM parameters. When prompted with contrastive prompts, LLMs can leverage these patterns to generate both a correct and a wrong answer. The "correct" answer is more likely to align with ground truth, as the model has learned to eliminate possible wrong answers. 2) Instruction tuning unlocks the abilities of LLMs to give correct and incorrect answers by fine-tuning on various natural language processing tasks including reasoning tasks~\cite{wei2021finetuned}. 3) RLHF fine-tunes LLMs using human feedback data, which offers relative judgments on the quality of answers. This feedback is valuable for enhancing the LLMs' capability to distinguish between correct and incorrect answers. 4) In CP, correct and wrong answers are returned by the LLM in a single output. The correct answers are generally distinct from the incorrect ones (as shown in Figure~\ref{fig:example_addsub}--\ref{fig:example_svamp}), thereby excluding the (mostly indeed) incorrect answers and reducing the probability of the correct answers being wrong. Before outputting the two answers, the LLM engages in "contrastive thinking" to determine which answer is correct and which is incorrect.


\section{Conclusion}
We propose CP, a template-based prompting approach for contrastive reasoning. Quantitative and qualitative results indicate that Zero-Shot-CP shows significant improvements across various reasoning tasks. Our method can seamlessly integrate with any prompting technique by incorporating a trigger sentence before the LLM provides answers. 
CP not only outperforms Zero-shot-CoT and Few-shot-CoT in most arithmetic and commonsense reasoning tasks, but also achieves comparable or even superior results when compared to state-of-the-art methods. 

\section{Limitations}
\label{limit}
Our work has some limitations and there is room for further exploration and improvement. Firstly, we have not yet validated the effectiveness of CP on smaller models such as Gemma-2B and Qwen1.5-0.5B. Secondly, we can further explore the combination of contrastive prompting with other prompting methods, such as X-of-thought approaches. Lastly, exploring the impact of contrastive prompting on LLM parameters and visualizing it would be an interesting future direction.


\bibliography{custom}

\newpage

\appendix
\section{Details of Experimental Setup}
\label{appendix:settings}
\subsection{Code, Prompts, Logs}



All code is available at~\url{https://github.com/yao8839836/cp}.

All prompts are available at~\url{https://github.com/yao8839836/cp/blob/master/main.py}.

Our experimental logs are available at~\url{https://github.com/yao8839836/cp/tree/master/log}.

\subsection{Prompts For Answer Extraction}
\label{appendix:answer_extract}
Table~\ref{tab:answer_extract} summarizes the answer extraction prompt for each task used for the CP experiments.

\subsection{Pre-training data examples}
\label{appendix:pretrain_data_examples}
For instance, many web pages and books provide correct and incorrect answers to math reasoning\footnote{\url{https://prek-math-te.stanford.edu/operations/analyzing-thinking-underlying-wrong-answers}}~\footnote{\url{https://mathmistakes.org/category/elementary-school/}}~\footnote{\url{https://www.gutenberg.org/ebooks/38769}} and common sense reasoning ~\footnote{\url{https://www.proprofs.com/quiz-school/story.php?title=common-sense-quiz_1}}~\footnote{\url{https://www.wikihow.com/Common-Sense-Quiz}} questions.

\section{Additional Experimental Results}
\label{appendix:extra_results}
\begin{figure*}[t]
  \centering
  \includegraphics[width = 0.80 \textwidth]{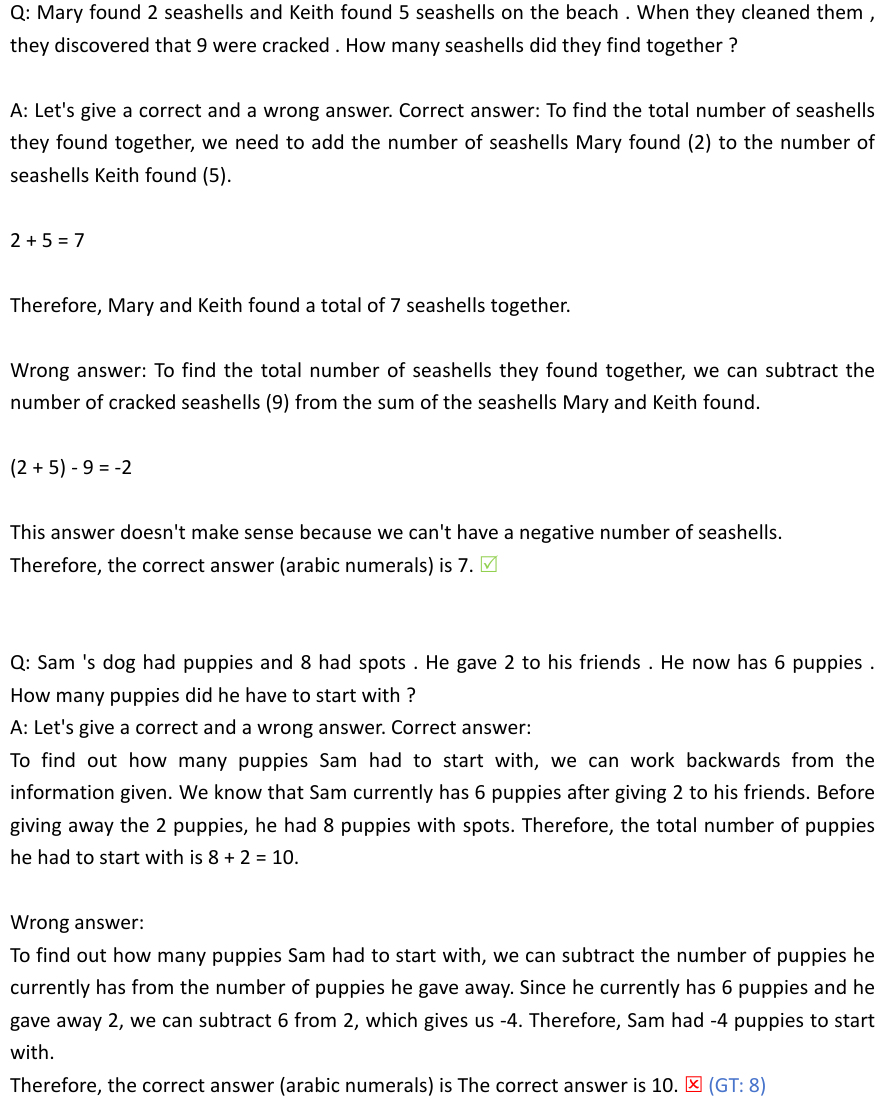}
  \caption{Example outputs by Zero-shot-CP for AddSub.}
  \label{fig:example_addsub}
\end{figure*}

\begin{figure*}[t]
  \centering
  \includegraphics[width = 0.80 \textwidth]{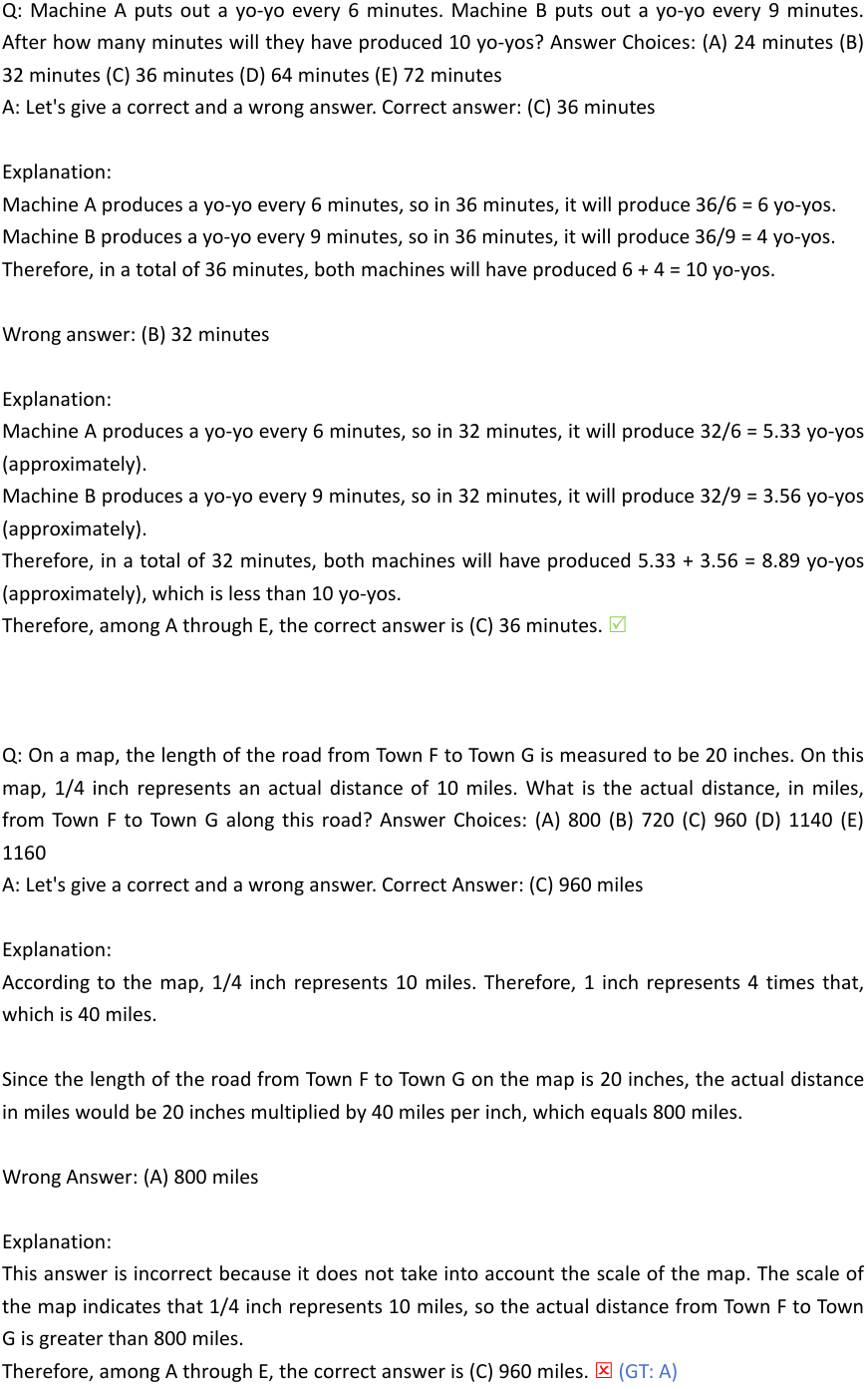}
  \caption{Example outputs by Zero-shot-CP for AQUA-ART.}
  \label{fig:example_aqua}
\end{figure*}

\begin{figure*}[t]
  \centering
  \includegraphics[width = 0.80 \textwidth]{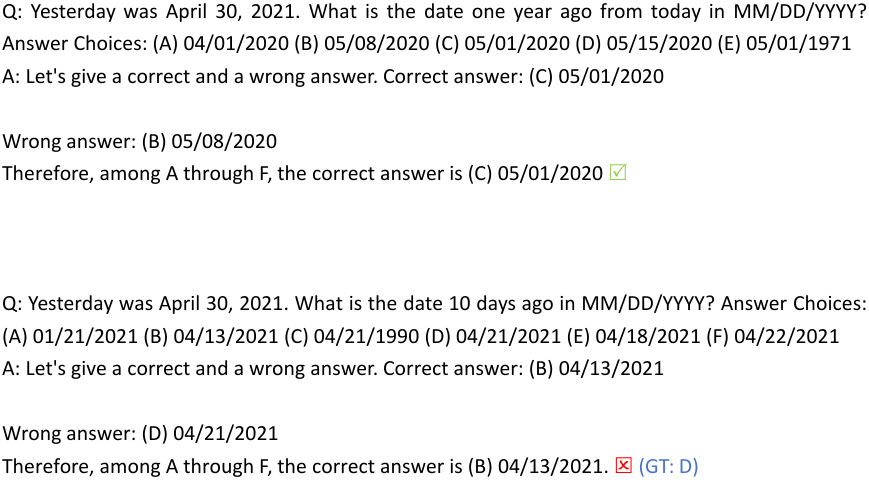}
  \caption{Example outputs by Zero-shot-CP for Date Understanding.}
  \label{fig:example_date}
\end{figure*}

\begin{figure*}[t]
  \centering
  \includegraphics[width = 0.80 \textwidth]{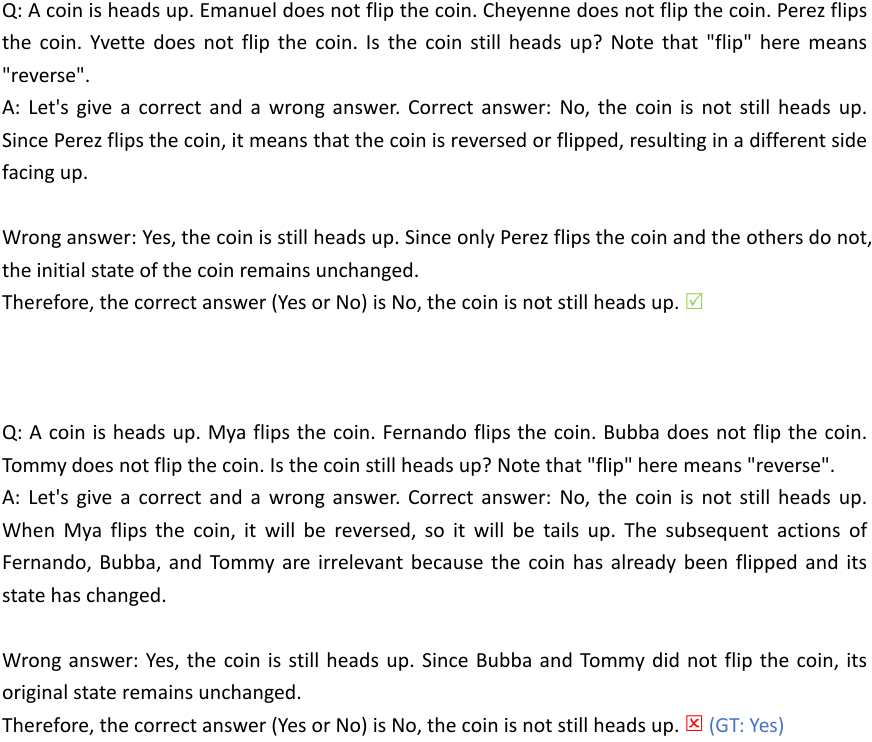}
  \caption{Example outputs by Zero-shot-CP for Coin Flip.}
  \label{fig:example_coinflip}
\end{figure*}

\begin{figure*}[t]
  \centering
  \includegraphics[width = 0.80 \textwidth]{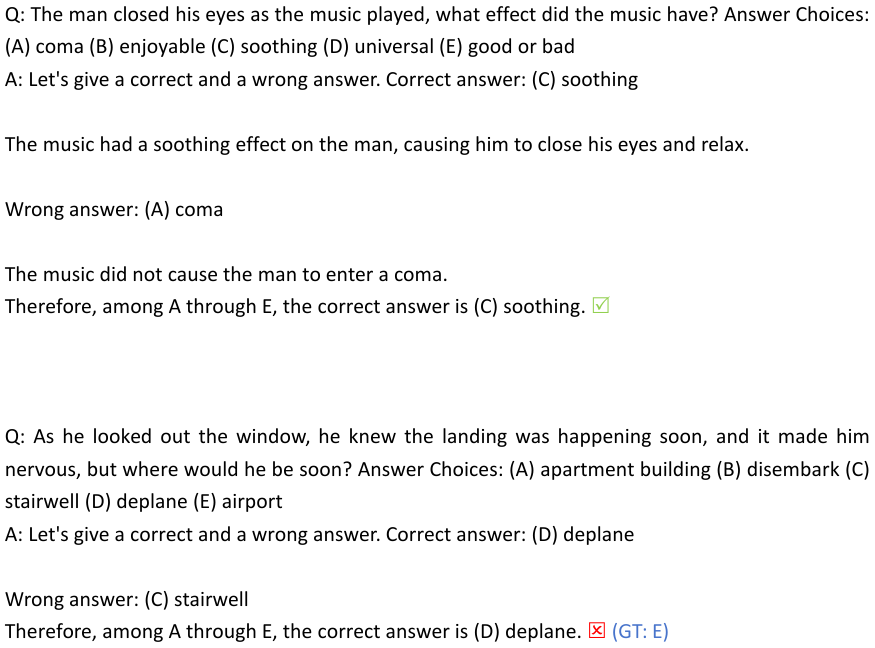}
  \caption{Example outputs by Zero-shot-CP for CommonsenseQA.}
  \label{fig:example_csqa}
\end{figure*}

\begin{figure*}[t]
  \centering
  \includegraphics[width = 0.80 \textwidth]{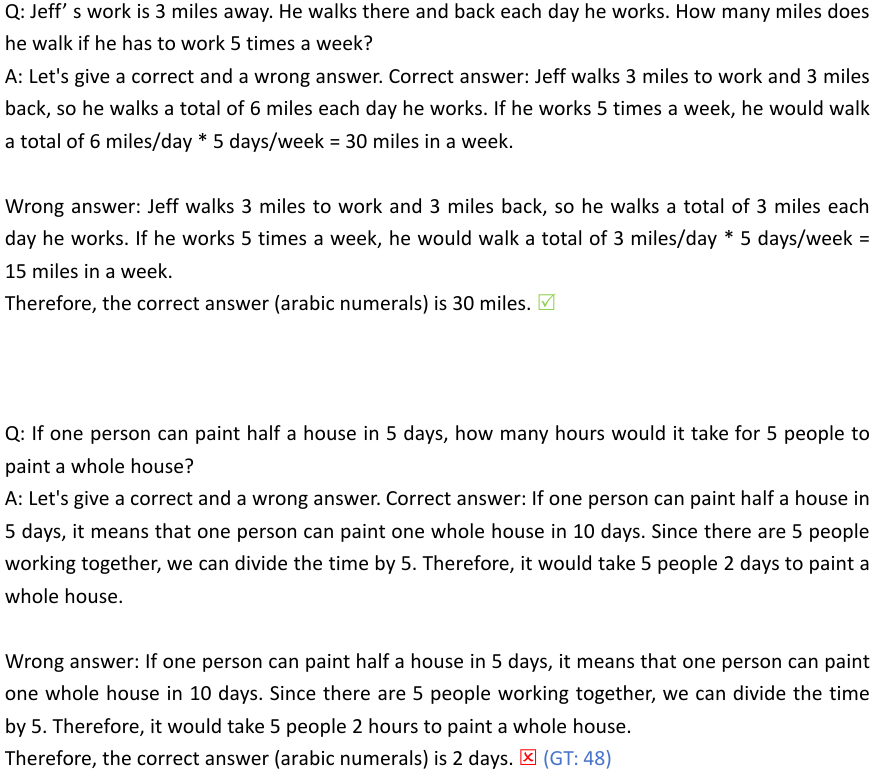}
  \caption{Example outputs by Zero-shot-CP for GSM8K.}
  \label{fig:example_gsm8k}
\end{figure*}

\begin{figure*}[t]
  \centering
  \includegraphics[width = 0.80 \textwidth]{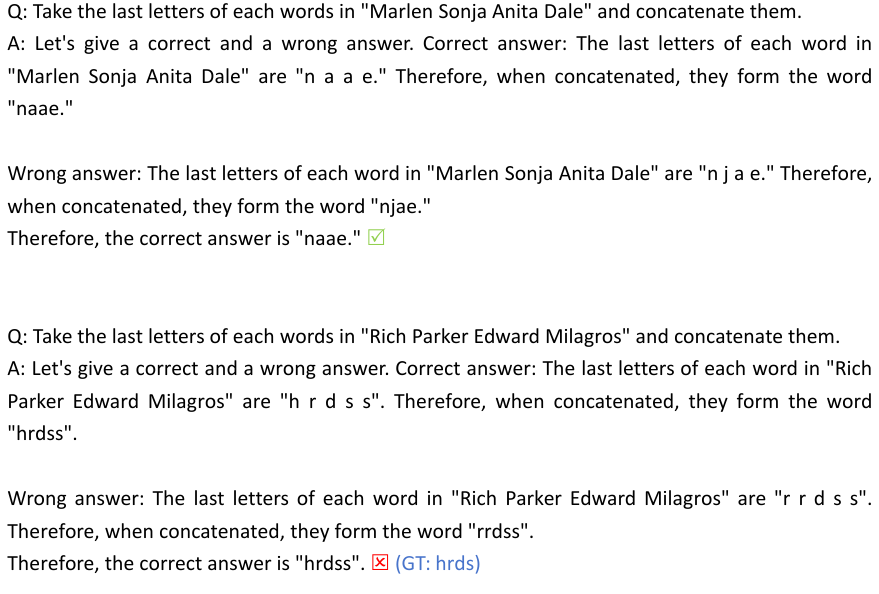}
  \caption{Example outputs by Zero-shot-CP for Last Letter Concatenation.}
  \label{fig:example_lastletter}
\end{figure*}

\begin{figure*}[t]
  \centering
  \includegraphics[width = 0.80 \textwidth]{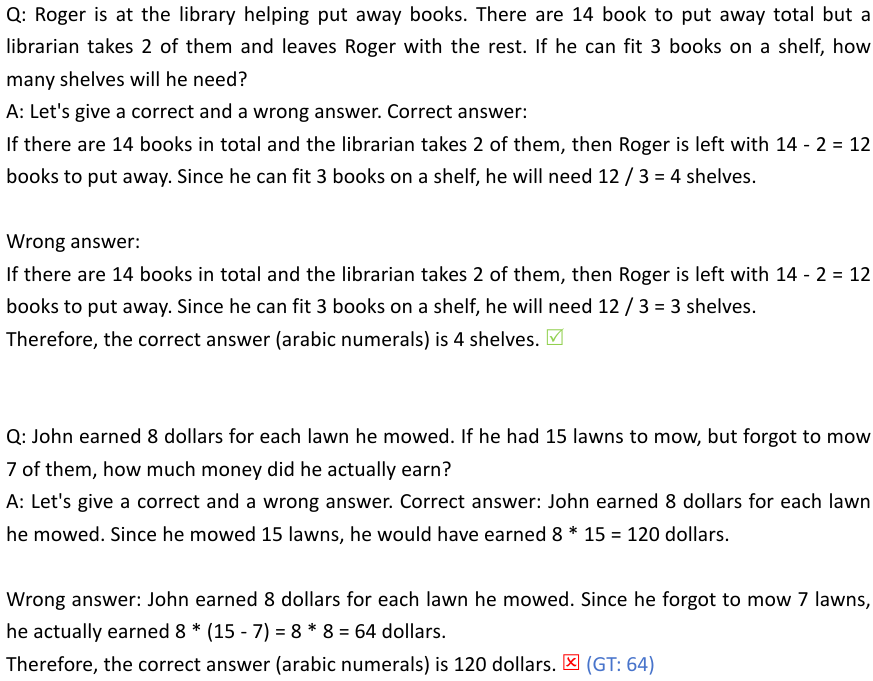}
  \caption{Example outputs by Zero-shot-CP for MultiArith.}
  \label{fig:example_multiarith}
\end{figure*}

\begin{figure*}[t]
  \centering
  \includegraphics[width = 0.80 \textwidth]{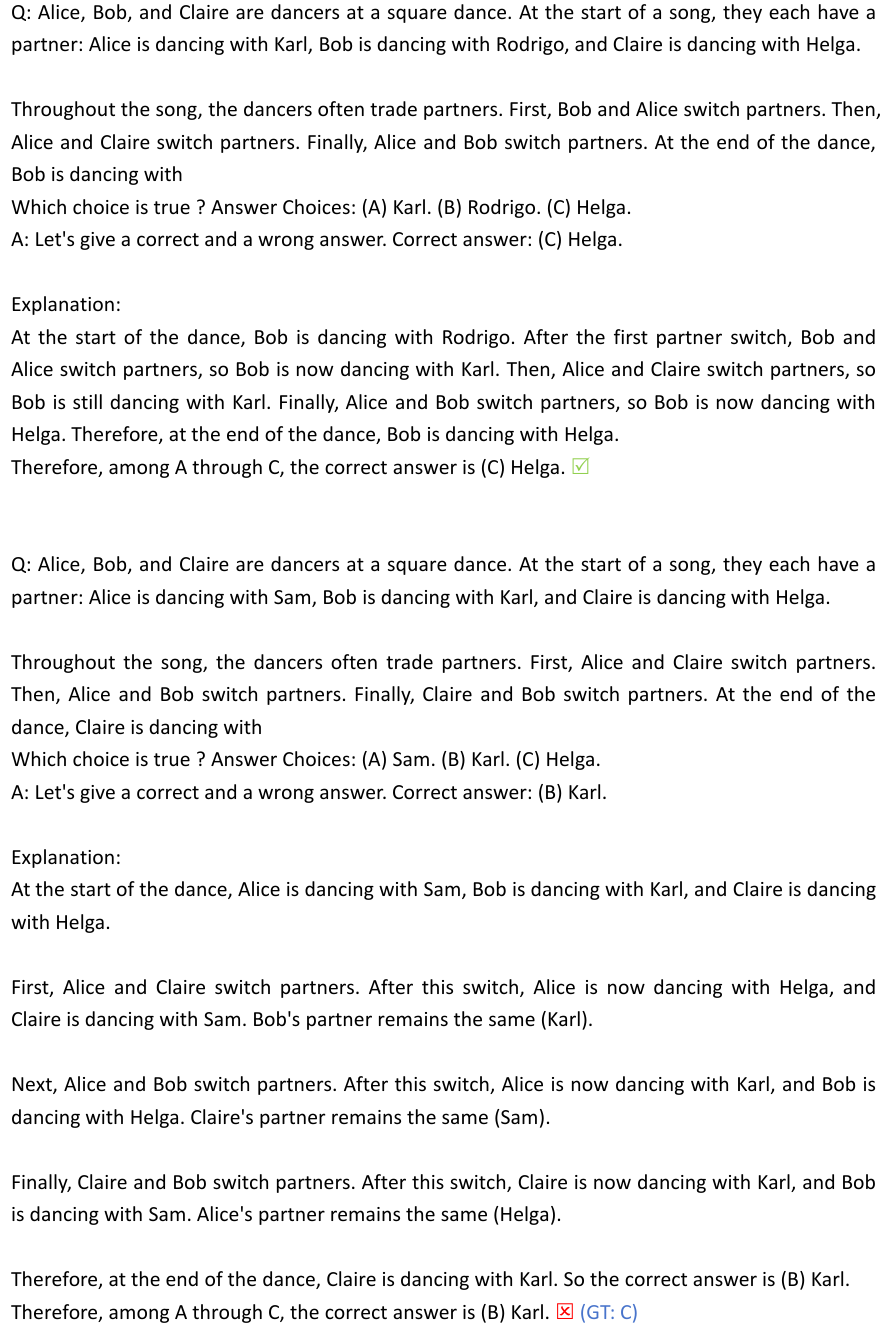}
  \caption{Example outputs by Zero-shot-CP for Tracking Shuffled Object.}
  \label{fig:example_object}
\end{figure*}

\begin{figure*}[t]
  \centering
  \includegraphics[width = 0.80 \textwidth]{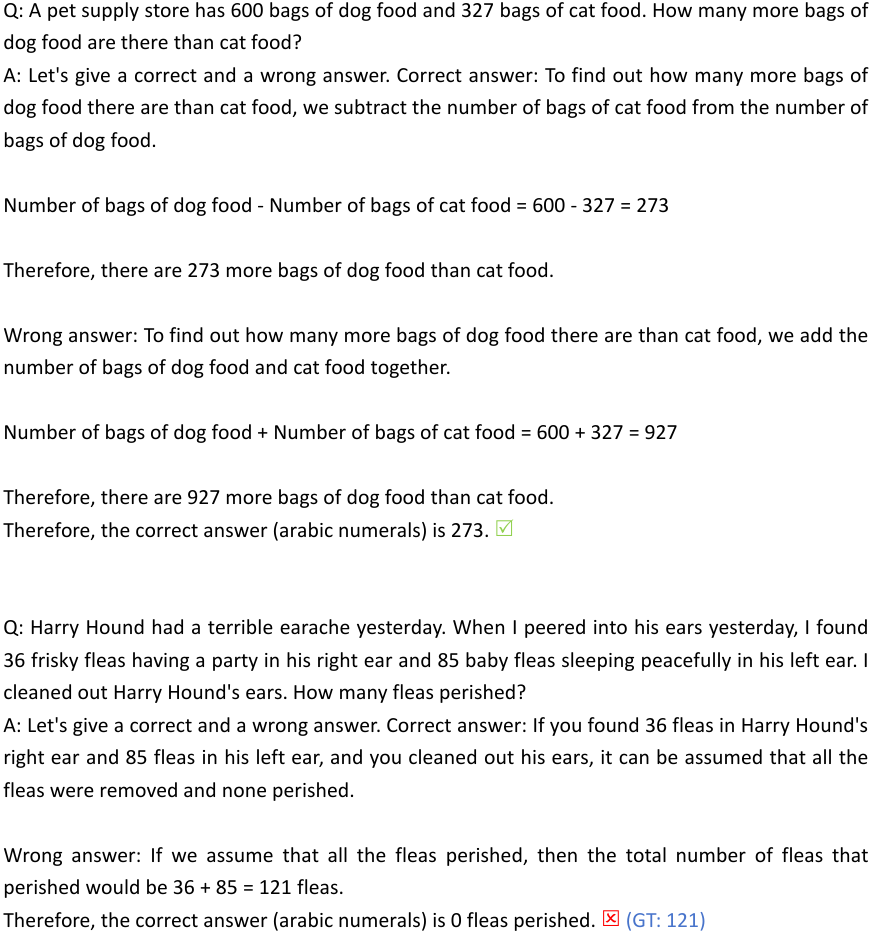}
  \caption{Example outputs by Zero-shot-CP for SingleEq.}
  \label{fig:example_singleeq}
\end{figure*}

\begin{figure*}[t]
  \centering
  \includegraphics[width = 0.80 \textwidth]{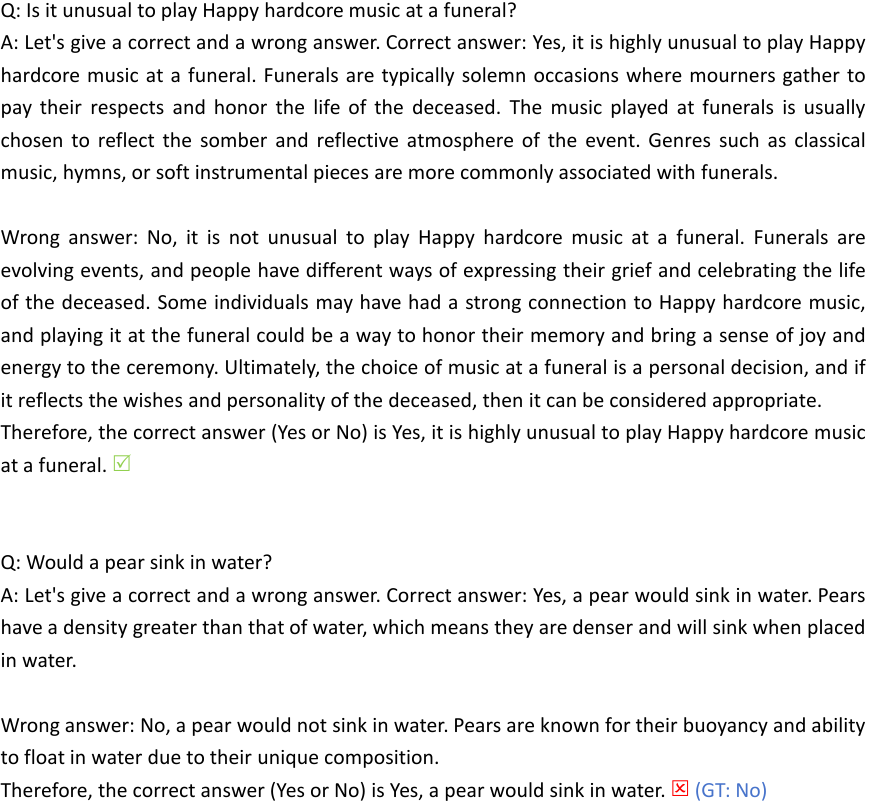}
  \caption{Example outputs by Zero-shot-CP for StrategyQA.}
  \label{fig:example_strategyqa}
\end{figure*}

\begin{figure*}[t]
  \centering
  \includegraphics[width = 0.80 \textwidth]{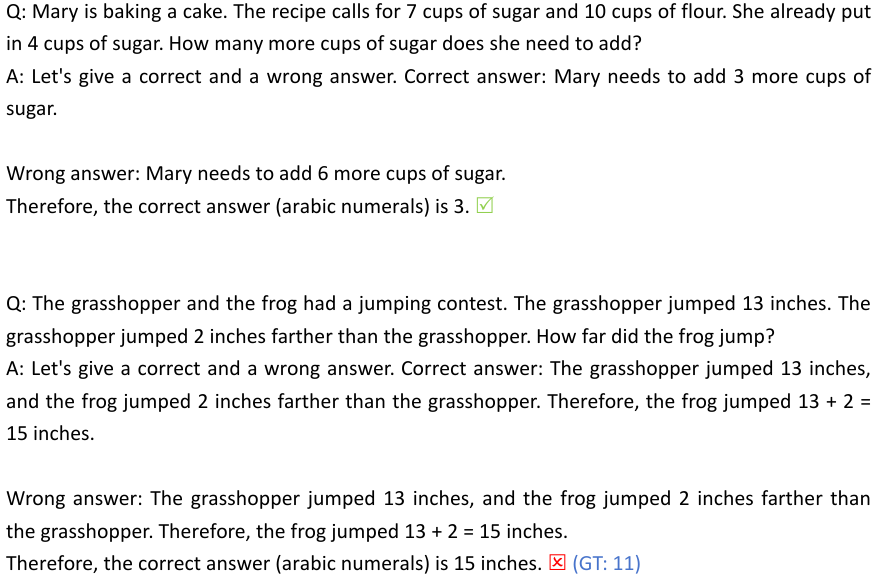}
  \caption{Example outputs by Zero-shot-CP for SVAMP.}
  \label{fig:example_svamp}
\end{figure*}

In this section, we provide a summary of additional example texts generated by Zero-shot-CP. GPT-3.5-Turbo is used as the model if not specified.  Table~\ref{tab:answer_extract_csqa_208st} illustrates example outputs of zero-shot prompting methods from CommonsenseQA. Figure~\ref{fig:example_addsub}--\ref{fig:example_svamp} show a positive example and a negative example of Zero-shot-CP on each dataset. "GT" in the figures means "Ground Truth".

The 240 examples, along with our annotations, can be accessed at the following link:~\url{https://anonymous.4open.science/r/cp-712E/results/zero_shot_cp_gpt4_240_examples_labeled.txt}.


Table~\ref{table:main_results_zero} presents the comparison of Zero-shot-CP with Zero-shot and Zero-shot-CoT on all 12 datasets using GPT-3.5-Turbo and GPT-4.

Table~\ref{tab:answer_extract_csqa_208st} presents an example question from CommonsenseQA and responses from different methods.

Table~\ref{table:results_openllm} presents the results of using various open-source LLMs: LLaMA3-8B, LLaMA3-70B, ChatGLM3-6B, and Qwen1.5-72B-Chat as base models.

Table~\ref{table:results_correct_only} presents the comparison of the results using "Let’s give a correct and a wrong answer." and "Let’s give a correct answer." prompts.

Table~\ref{table:240_examples} provides the categorization and counts of these 120 solved problems and 120 unsolved problems.

\begin{figure*}[t]
  \centering
  \includegraphics[width = 0.80 \textwidth]{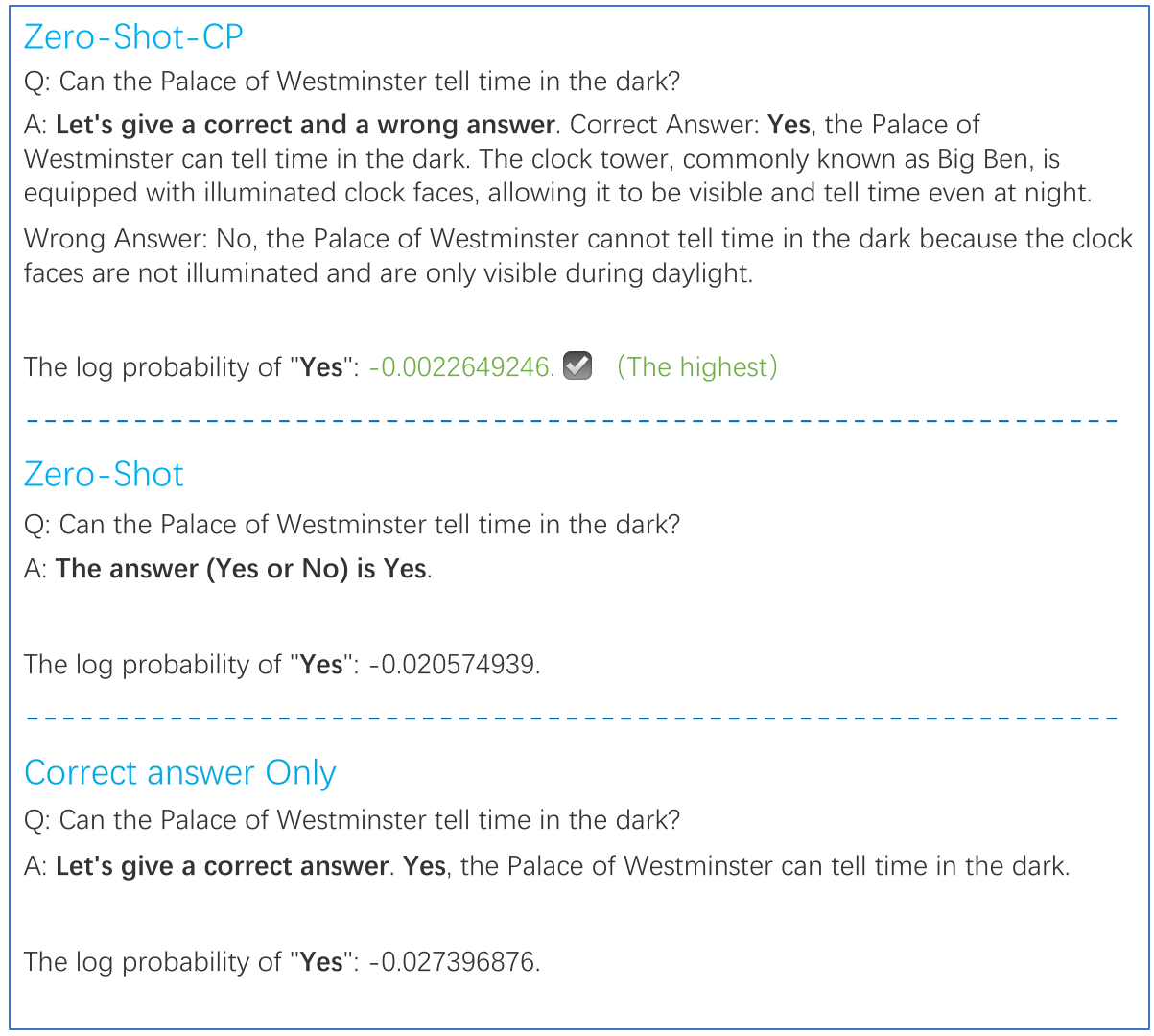}
  \caption{By setting the logprobs (log probabilities) parameter of the OpenAI API (using GPT-4), we printed the token output probabilities for different prompts. We provide an example in StrategyQA. The ground truth is "Yes". Note that the higher the logprobs value, the greater the probability. Zero-shot-CP makes GPT-4 more confident in the answer compared to Zero-shot and "Let's give a correct answer.".}
  \label{fig:token_prob}
\end{figure*}

In Figure~\ref{fig:token_prob}, we printed the
token output probabilities for different prompts. We provide an example in StrategyQA.

\begin{table*}[t]
  \footnotesize
  \centering
  \begin{tabular}{lllllll}
    \toprule
    &\multicolumn{6}{c}{Arithmetic}                   \\
    \cmidrule(r){2-7}
    Method     & SingleEq     & AddSub & MultiArith & GSM8K & AQUA &  SVAMP\\
    \midrule
    Zero-shot & 90.6/81.7 & \textbf{92.4}/82.8 & 96.5/61.2& 35.9/14.3& 41.3/29.9 & 86.4/69.7\\
    Zero-shot-CoT & 91.7/91.1 & 89.6/86.6 & 97.7/94.8& \textbf{90.9}/\textbf{75.1}& \textbf{70.1}/\textbf{55.9} & 90.4/81.9\\
    Zero-shot-CP & \textbf{91.7}/\textbf{91.7} &  91.6/\textbf{90.6} & \textbf{97.8}/\textbf{95.2}& 88.8/73.2& 62.2/40.2 & \textbf{91.5}/\textbf{83.2}\\
    \toprule
    &\multicolumn{2}{c}{Common Sense}    & \multicolumn{2}{c}{Other Reasoning Tasks}  & \multicolumn{2}{c}{Symbolic Reasoning}                \\
    \cmidrule(r){2-7}
        & Common    & Strategy & Date & Shuffled & Last Letter &  Coin Flip\\
        & SenseQA    & QA & Understand & Objects &  (4 words) &   (4 times)\\
    \midrule
    Zero-shot & 82.9/71.3 & 64.8/65.0 & 73.2/40.4& 40.7/33.9& 5.0/4.2 & 36.6/49.6\\
    Zero-shot-CoT & 78.3/67.8 & 69.8/60.9 & \textbf{79.4}/\textbf{62.1}& \textbf{93.1}/\textbf{73.1} & \textbf{90.2}/\textbf{88.0} & \textbf{98.6}/\textbf{94.0}\\
    Zero-shot-CP & \textbf{83.5}/\textbf{73.9} &  \textbf{73.4}/\textbf{67.3} & 71.5/51.8& 44.4/51.5& 23.4/41.8 & 33.2/56.8\\
    \bottomrule
  \end{tabular}
    \caption{Accuracy (in percentage) comparison of Zero-shot-CP with Zero-shot and Zero-shot-CoT on each dataset. The values on the left-hand side of each task represent the results obtained using GPT-4, while the values on the right-hand side represent the results obtained using GPT-3.5-Turbo.}
    \label{table:main_results_zero}
\end{table*}

\begin{table*}[t]
\scriptsize

  \centering
  \begin{tabular}{l|l}
    \toprule
    \multirow{2}{*}{Question}   & Q: A bald eagle is likely to be found on what kind of work? \\
                  &~~~~~Answer Choices: (A) rural area (B) book (C) canada (D) painting (E) aviary\\ 
    \midrule
    \multirow{4}{*}{Zero-shot} & A: Among A through E, the answer is {\color{red}(E) aviary}. However, it should be noted that\\ 
    &~~~~~bald eagles can also be found in rural areas (A), in Canada (C), and depicted in books (B) and paintings (D).  \\
    &~~~~~The question seems to be asking where a bald eagle is likely to be found physically, and an aviary is a place\\
    &~~~~~where birds are kept.\\
    \midrule
     \multirow{7}{*}{Zero-shot-CoT} & A: Let's think step by step. A bald eagle is a bird, so it could be found in an aviary (E), which is a place where\\
    &~~~~~birds are kept. It could also be found in a rural area (A), as these birds often live in areas with less human\\
    &~~~~~activity. A bald eagle could be depicted in a book (B) or a painting (D), as it is a popular symbol in various\\
    &~~~~~cultures, especially in the United States. Lastly, bald eagles are also found in Canada (C), as their habitat\\
    &~~~~~ranges across most of North America. Therefore, all of the options could be correct depending on the\\
    &~~~~~context of the question. \\
    &~~~~~Therefore, among A through E, the answer is {\color{red}(A) rural area, (B) book, (C) Canada, (D) painting, (E) aviary.}\\
    \midrule
    \multirow{4}{*}{Zero-shot-CP} & A: Let's give a correct and a wrong answer. \textbf{Correct Answer}: (D) painting\\    
    & \\
    &~~~~~\textbf{Wrong Answer}: (A) rural area\\
    &~~~~~Therefore, among A through E, the correct answer is {\color{green}(D) painting}\\    
    \midrule    
  Ground Truth  &~~~~~D  \\  
    \bottomrule
  \end{tabular}
  \caption{An example question from CommonsenseQA and responses from Zero-shot, Zero-shot-CoT, and Zero-shot-CP. We use GPT-4 as the base model.}
  \label{tab:answer_extract_csqa_208st}
\end{table*}

\begin{table*}
  \footnotesize
  \centering
  \begin{tabular}{lll}
    \toprule
    No     & Task     & CP answer extraction prompts \\
    \midrule
    1 & SingleEq  & Therefore, the correct answer (arabic numerals) is     \\
    2     & AddSub & Therefore, the correct answer (arabic numerals) is      \\
    3     & MultiArith       &  Therefore, the correct answer (arabic numerals) is  \\
    4     & GSM8K       &  Therefore, the correct answer (arabic numerals) is  \\
    5     & AQUA-RAT       &  Therefore, among A through E, the correct answer is  \\
    6     & SVAMP       &  Therefore, the correct answer (arabic numerals) is  \\
    7     & CommonsenseQA      &  Therefore, among A through E, the correct answer is \\
    8     & StrategyQA       &  Therefore, the correct answer (Yes or No) is \\
    9     & Date Understanding      & Therefore, among A through F, the correct answer is  \\
    10     & Shuffled Objects     & Therefore, among A through C, the correct answer is  \\
    11     & Last Letters     & Therefore, the correct answer is  \\
    12     & Coin Flip     & Therefore, the correct answer (Yes or No) is  \\
    \bottomrule
  \end{tabular}
  \caption{Answer extraction prompts used in our CP experiments.}
  \label{tab:answer_extract}
\end{table*}

\begin{table*}[t]

  \footnotesize
  \centering
  \begin{tabular}{lccccc}
    \toprule
        & MultiArith    & GSM8K & StrategyQA & AQUA & SVAMP \\
    \midrule
    \textit{LLaMA3-8B}&     &  &  &  &  \\
    Zero-shot &  31.0 & 38.1 & --& --& 52.8\\
    Zero-shot-CP & \textbf{57.3} & \textbf{54.9} & --& --& \textbf{61.4}\\
    \midrule
    \textit{LLaMA3-70B}&     &  &  &  &  \\
    Zero-shot &  86.5 & 63.7 & 54.5& 38.2& --\\
    Zero-shot-CP & \textbf{97.0} & \textbf{66.1} & \textbf{57.5}& \textbf{62.2}& --\\
    \midrule
    \textit{ChatGLM3-6B}&     &  &  &  &  \\
    Zero-shot &  5.3 & 4.3 & --& --& --\\
    Zero-shot-CP & \textbf{67.0} & \textbf{40.0} & --& --& --\\
    \midrule
    \textit{Qwen1.5-72B-Chat}&     &  &  &  &  \\
    Zero-shot &  54.7 & 19.3 & 71.2& 31.1& 65.2\\
    Zero-shot-CP & \textbf{75.5} & \textbf{52.1} & \textbf{73.5}& \textbf{45.3}& \textbf{77.4}\\
    \bottomrule
  \end{tabular}
  \caption{Accuracy (in percentage) comparison of Zero-shot-CP with Zero-shot using open LLMs.}
  \label{table:results_openllm}
\end{table*}

\begin{table*}[t]

  \footnotesize
  \centering
  \begin{tabular}{lcc}
    \toprule
        & CommonsenseQA    & StrategyQA  \\
    \midrule
    \textit{GPT-3.5-Turbo}&     &   \\
    Let's give a correct answer. &  73.1 & 64.4  \\
    Let's give a correct and a wrong answer. & \textbf{73.9} & \textbf{67.3}  \\
    \midrule
    \textit{GPT-4}&     &   \\
    Let's give a correct answer. &  82.3 & 71.8 \\
    Let's give a correct and a wrong answer. & \textbf{83.5} & \textbf{73.4} \\
    \bottomrule
  \end{tabular}
  \caption{Comparison of the results using "Let's give a correct and a wrong answer." and "Let's give a correct answer." prompts.}
  \label{table:results_correct_only}
\end{table*}

  \begin{table*}[t]
  \scriptsize
  \centering
  \begin{tabular}{lc}
    \toprule
      \textbf{Category}  & \textbf{\# Examples}     \\
    \midrule
    The given "correct" answer is the GT, and the given "wrong" answer is indeed incorrect. &  112\\
    The given "correct" answer is the GT, and the given "wrong" answer is also the GT. &  4\\
    The given "correct" answer is the GT, no "wrong" answer is given. &  4\\
    \midrule
    The given "correct" answer is incorrect, and the given "wrong" answer is the GT. &  23\\
    The given "correct" answer is incorrect, and the given "wrong" answer is also incorrect. &  91\\
    The given "correct" answer is incorrect, no "wrong" answer is given. &  6\\
    \bottomrule
  \end{tabular}
  \caption{Categorization results of Zero-shot-CP output (with GPT-4) on 240 problems. We manually annotated 10 solved problems and 10 unsolved problems for each of the 12 datasets. GT means Ground Truth. See Appendix~\ref{appendix:extra_results} for the link of the examples.}
  \label{table:240_examples}
\end{table*}

\end{document}